%% file: acl.tex
\pdfoutput=1

\documentclass[11pt]{article}

\usepackage[]{acl}

\usepackage{times}
\usepackage{latexsym}
\usepackage{graphicx}
\usepackage{multirow}
\usepackage{tabularx}
\usepackage{booktabs} 
\usepackage{chngpage}
\usepackage{tabularx}

\usepackage[T1]{fontenc}

\usepackage[utf8]{inputenc}

\usepackage{microtype}

\def\SM{{\mathcal S}}

\usepackage{array}
\usepackage{amsmath}
\usepackage{bm}

\usepackage{url}
\usepackage{titlesec}
\usepackage{caption}
\usepackage{subcaption}
\usepackage{url}
\usepackage{inconsolata}
\usepackage[shortlabels]{enumitem}

\usepackage{color} 
\usepackage{url}

\title{Multilingual Word Sense Disambiguation with \\ Unified Sense Representation}

\author{{\bf Ying Su}$^1$, {\bf Hongming Zhang}$^{1,2}$, {\bf Yangqiu Song}$^1$,
{\bf Tong Zhang}$^1$ \\
$^{1}$HKUST, $^{2}$Tencent AI lab, Seattle \\
 \texttt{ysuay@connect.ust.hk},
 \texttt{\{hzhangal,yqsong\}@cse.ust.hk}, \\ \texttt{tongzhang@ust.hk}
\\
}

\begin{document}
\maketitle

\input{files/1-abstract}
\input{files/2-introduction}

\input{files/3-related-works}
\input{files/4-approach}

\input{files/5-experiments}

\input{files/6-results}

\input{files/7-conclusion}

\bibliography{anthology,custom}
\bibliographystyle{acl_natbib}

\appendix



\end{document}

%% file: files/1-abstract.tex
\begin{abstract}
As a key natural language processing (NLP) task, word sense disambiguation (WSD) evaluates how well NLP models can understand the lexical semantics of words under specific contexts.
Benefited from the large-scale annotation, current WSD systems have achieved impressive performances in English by combining supervised learning with lexical knowledge.
However, such success is hard to be replicated in other languages, where we only have limited annotations.
In this paper, based on the multilingual lexicon BabelNet describing the same set of concepts across languages, we propose building knowledge and supervised-based Multilingual Word Sense Disambiguation (MWSD) systems.
We build unified sense representations for multiple languages and address the annotation scarcity problem for MWSD by transferring annotations from rich-sourced languages to poorer ones.
With the unified sense representations, annotations from multiple languages can be jointly trained to benefit the MWSD tasks. 
Evaluations of SemEval-13 and SemEval-15 datasets demonstrate the effectiveness of our methodology.

\end{abstract}

%% file: files/2-introduction.tex
\section{Introduction}
As a critical natural language understanding task, word sense disambiguation (WSD) aims at classifying words into pre-defined senses.
With such a disambiguation process, machines can understand the precise meanings of words.
Previous researches have demonstrated that a sound WSD system could benefit many downstream NLP tasks, such as machine translation \cite{pu2018integrating, liu2018handling} and information extraction \cite{bovi2015knowledge}.

\begin{figure}[t]
\centering
\includegraphics[width=\linewidth]{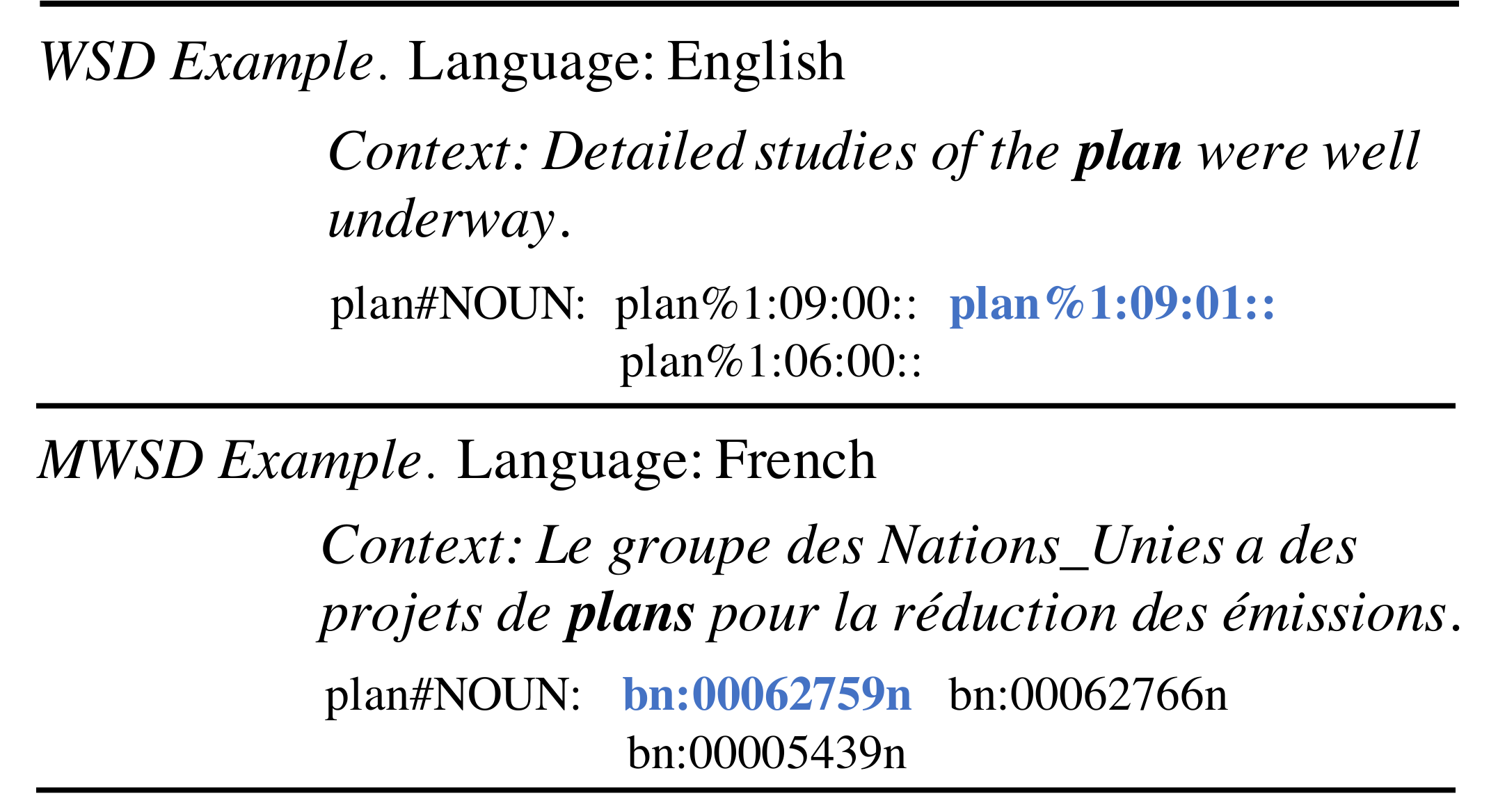}
\caption{Examples of WSD and Multilingual WSD (MWSD) task. The target words are indicated with the bold font in contexts. Candidate sense keys are listed below each context, and the one in blue is the correct sense.}
\label{fig:intro-example}
\end{figure}

Existing researches on word sense disambiguation mostly focus on English only.
By leveraging lexical knowledge such as gloss \cite{iacobacci2016embeddings, luo2018incorporating,  huang2019glossbert, blevins2020moving} or graph structure \cite{banerjee2003extended, kumar2019zero, bevilacqua2020breaking} and supervised training over large-scale annotations, these models have achieved impressive performance on the standard English WSD task.
However, though the English WSD task \cite{raganato2017word} and multilingual WSD (MWSD) task \cite{navigli2013semeval, moro2015semeval} are of the same form as shown in Figure \ref{fig:intro-example}, this progress can not be easily applied across languages as the paucity of annotated training data and immense labor in handling diverse lexical knowledge of multiple languages separately. 

\begin{figure}[t]
\centering
\includegraphics[scale=0.47, trim={0.0cm 0 0.0cm 0}]{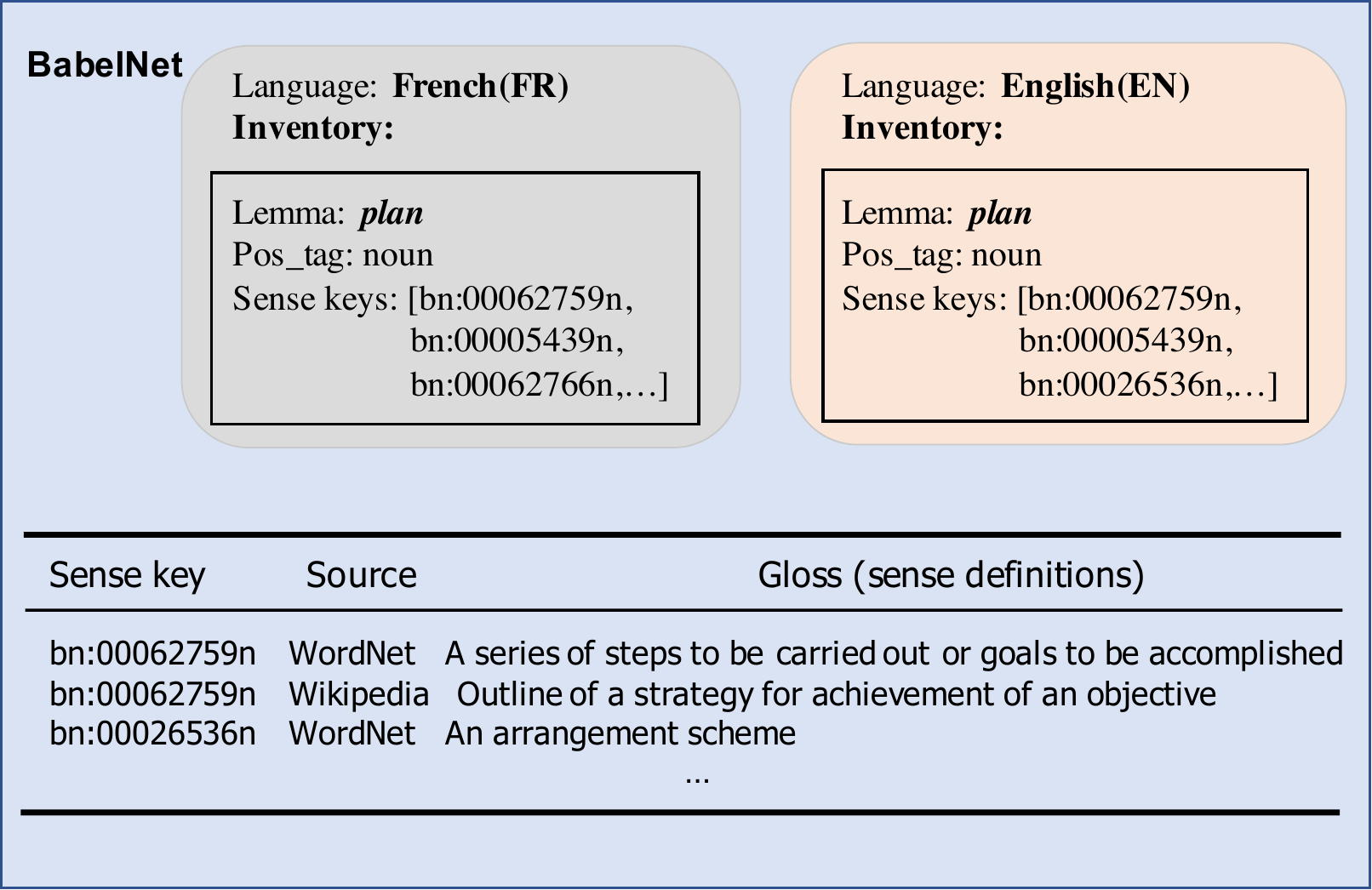}
\caption{BabelNet contains inventories for multiple languages. Each word in a language has several senses, and different words across languages may share the same senses. For each sense across languages, glosses from various sources such as WordNet and Wikipedia are collected to describe its meaning.}
\label{fig:intro-babel}
\end{figure}

BabelNet \cite{NAVIGLI2012217} is a multilingual semantic lexicon and contains a set of multilingually lexicalized concepts. Similar to WordNet \cite{miller1998wordnet}, a Babel synset defines a concept shared by a group of words across languages with the same meaning. Based on the multilingual lexicon source BabelNet, we propose to build multilingual word sense disambiguation systems by inducing lexical knowledge and annotations from rich sourced language (e.g., English) to scarce sourced ones. First, as defined in BabelNet, words in each synset have the same sense, and the sense is described by lexical knowledge gloss despite the language forms. An example is shown in Figure \ref{fig:intro-babel}. The knowledge can be injected into supervised MWSD systems. Second, the annotations acquired from rich sourced languages through machine translation and alignment tools, can be used as weak supervision. By utilizing the lexical knowledge and weak annotations, we can build a decent MWSD system for scarce sourced languages without further human effort.

To summarize, the contributions of this paper are two-fold:
(1) We propose to build an MWSD system mBERT-UNI for multiple languages with transferred annotations from rich sourced languages and unified synsets with lexical knowledge, addressing the data paucity problem on the MWSD task;
(2) Our system can be easily combined with other data generation efforts such as MuLaN ~\cite{barba2020mulan}, further boosting the system performance.
Experiments results on benchmark SemEval-13 \cite{navigli2013semeval} and SemEval-15 \cite{moro2015semeval} demonstrate the effectiveness of our methodology. Our code is open-resourced\footnote{https://github.com/suytingwan/multilingual-WSD}.

%% file: files/3-related-works.tex
\section{Related Work}
This section introduces previous efforts on multilingual word sense disambiguation, which can be categorized into two streams: data-driven systems and knowledge-based systems.

\subsection{Data-driven Systems}

In the last decades, many efforts in the field of multilingual word sense disambiguation have been devoted to mitigating the knowledge acquisition bottleneck problem \cite{gale1992method, pasini2020knowledge}, which is hard to acquire sense-annotated corpora for multiple languages. 
To mitigate the paucity of annotations, many researchers have focused on automatically creating high-quality, sense-annotated training corpora \cite{pasini2020train}. OMSIT \cite{taghipour2015one} proposed a semi-automatic approach to acquire one million training instances from MultiUN dataset \cite{eisele2010multiun}. OneSec \cite{scarlini2019just} proposed to generate multilingual sense-annotated datasets on a large scale by mapping Wikipedia categories to word senses. MuLaN \cite{barba2020mulan} utilized contextualized word embeddings to transfer sense annotations from labeled datasets SemCor \cite {miller1993semantic} and WNG \cite{langone2004annotating} to the unlabeled corpus from Wikipedia across languages. \citet{hauer2021semi} proposed a label propagation approach for constructing multilingual sense-annotated corpora by machine translation. XL-WSD \cite{pasini2021xl} further enriches the annotations across 18 languages from six different linguistic families. 
Similar to \citet{hauer2021semi}, our automatic corpora generation method takes advantage of machine translation and alignment tools, while it is easy and feasible to use without additional resources. Moreover, we also induce lexical knowledge in building sense representations.

\subsection{Knowledge-based Systems}
Besides annotated corpora, lexical knowledge such as sense inventories is another key component in word sense disambiguation systems. Lexical knowledge sources such as WordNet and BabelNet provide rich lexical knowledge, e.g., gloss or graph structure. Such knowledge has been exploited and shows decent performance in many supervised systems \cite{kumar2019zero, loureiro2019language, scarlini2020sensembert, blevins2020moving}. Readers can refer to \cite{bevilacqua2021recent} for more details.



In this paper, we aim to induce lexical knowledge into MWSD systems. Based on the synsets and lexical knowledge in multilingual lexicon BabelNet, we propose to build unified sense representations that can be shared across languages. The sense representations can be incorporated into supervised systems to improve the performance of MWSD tasks.


%% file: files/4-approach.tex
\begin{figure*}[t]
\centering
\includegraphics[scale=0.46]{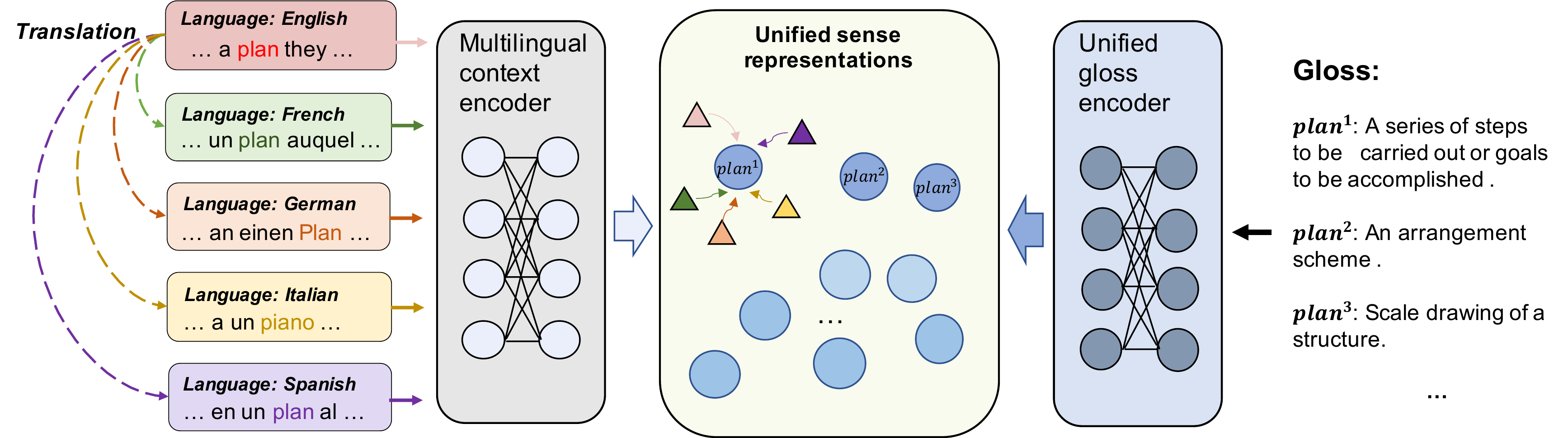}
\caption{Architecture Overview of the mBERT-UNI model. We apply multilingual BERT for the encoders. The context encoder takes the multilingual context as input and generates representations for the target words. The gloss encoder takes the glosses as input and generates unified sense representations. The similarity scores between target word embedding and representation of candidate senses are calculated for each language. The sense with the highest score is the predicted sense by the model.}
\label{fig:overview}\vspace{-1em}
\end{figure*}

\section{Approach}

In this section, we first present the formal definition of the multilingual WSD task and used notations. 
After that, we present the details of the proposed system mBERT-UNI, a supervised framework incorporating lexical unified representation space for the MWSD task. 
From the overview in Figure~\ref{fig:overview}, we can see that mBERT-UNI can be decomposed into four parts: 
(1) To address the data paucity issue, we first translate the annotated English corpus SemCor into other languages and use alignment tool to generate sense annotations; (2) A context encoder encodes the target words in multilingual context; 
(3) A gloss encoder encodes the glosses to produce unified sense representations;
(4) The annotated corpus in several languages and unified sense representations are bound with a joint training setting.

\subsection{Task Description and Notations}
In the multilingual setting, the WSD task is to disambiguate the senses for a sequence of words $\left\{w_{1}, \cdots, w_{m}\right\}$ in a sentence $S$. The sentences come from various languages $L \in \left\{L_{1}, L_{2}, ...,L_{n}\right\} $. For each word $w$, the goal is to map it to a pre-defined sense $s \in \SM_m$, where $\SM_m = \{s_1, s_2, ..., s_k\}$ is the set of pre-defined candidate senses for $w$. The meaning of each sense is defined by the gloss. The candidate senses have a corresponding gloss set defined as $G=\{g_1, g_2, ..., g_k\}$. For the MWSD task, multiple languages have different inventories but share the same set of synsets and glosses as defined in BabelNet \cite{NAVIGLI2012217}.

\subsection{Multilingual Corpora Preparation}

\begin{figure*}[t]
\centering
\includegraphics[scale=0.5]{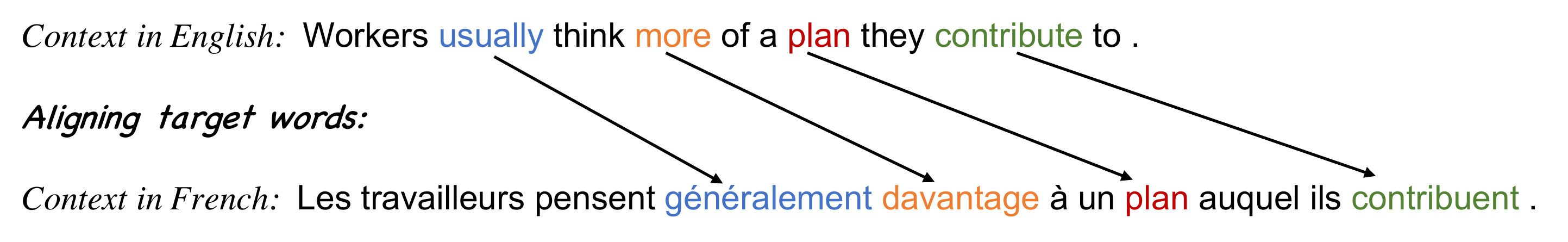}
\caption{Example of training corpora preparation. The training corpora are first translated into multiple languages by the google translation tool. Then FastAlign tool is applied to find the alignment of labeled target words in English context to other languages. The matched aligned words in different languages are shown in the same color.}
\label{fig:trans_align}
\end{figure*}

We use machine translation and alignment tools to acquire annotated training data for multiple languages. Machine translation has been developed for decades and has achieved remarkable progress \cite{wu2016google, tiedemann2020opus}. Following \cite{luan2020improving}, we use google translation to acquire parallel training corpora from English to other languages. Specifically, we translate SemCor \cite {miller1993semantic} into the target languages with the Google translation tool\footnote{https://translate.google.com} \cite{wu2016google}. 

There are many research alignment methods to acquire aligned words across languages based on parallel corpora \cite{dyer2013simple, ostling2016efficient, luan2020improving, dou2021word}. We use the early FastAlign tool\footnote{https://github.com/clab/fast\_align}  \cite{dyer2013simple} to align the words across languages for simplicity. Through the process, we propagate the annotations from English to multiple languages. The weak supervision signal in the transferred annotations can be further utilized in supervised systems.

An example of the context translation alignment and sense mapping between English and French is shown in Figure \ref{fig:trans_align}. Note that our method is language-independent and thus can be applied to many languages. For evaluating the MWSD system on SemEval-13 and SemEval-15, we apply the method to four languages, German (DE), French (FR), Spanish (ES), and Italian (IT).

\subsection{Model Overview}

The MWSD system mainly consists of an mBERT-UNI model, which is built upon the biencoder model for the English WSD task \cite{blevins2020moving}, where one encoder for encoding multilingual context and the other encoder for encoding unified gloss knowledge. The difference is that we apply the model to construct unified representations which can be used across languages. Both encoders apply the multilingual BERT (mBERT) transformer. mBERT \cite{kenton2019bert} is trained on 104 languages and is commonly used for cross-lingual semantic representation.

The model first extracts representations of context word and candidate sense representations. For a target word $w$ in the language $L$, the context encoder generates its representation as $e^{L}_{w}$. Specifically, it is the average pooling of target word tokens in the output. 

For sense representations, the gloss set of the corresponding target word is fed as input to the gloss encoder. Though different languages vary in the form of contexts, they share the synsets which can be described by gloss knowledge in a single language, e.g., English. The hidden state of [CLS] token in the gloss encoder output is the vector representation of the gloss. The representation of the candidate gloss set is $\left\{e_{g_{1}}, \cdots, e_{g_{k}}\right\}$. 


The similarity scores between embedding of target word and embeddings of its candidate sense set are calculated by the dot product:
$$score^{L}(w, g_{i}) = e^{L}_{w} \cdot e_{g_{i}}, i \in [1, k]. $$ 
The candidate sense with the highest score is the predicted sense produced by the system. 

The system is trained with cross-entropy loss over the scores after a softmax activation under a supervised paradigm:

$$
p_{i}=\frac{\exp ({score^{L}(w, g_{i})})}{\sum_{i=1}^{k} \exp ( score^{L}(w, g_{i}))},
$$
$$
loss^{L}(w) =- \sum_{i=1}^{k}\left[ y_{i} \log \left(p_{i}\right)\right],
$$
where $y_{i}$ is 1 if the $ith$ sense is the correct sense otherwise $y_{i}$ is 0.   \\

\noindent





\textbf{Joint Training Setting}. As annotations are scarce for low-resourced languages, we further design the joint language setting to see if the unified sense representation can connect annotations across languages to benefit the MWSD task. Under the joint language training setting, the inputs to the context encoder can be from different sources and languages. In contrast, the gloss encoder still generates representation for the Babel synsets.

%% file: files/5-experiments.tex
\section{Experiment Setup}

In this section, we introduce the experiment details including the evaluation dataset, the evaluation metric, training corpora, baseline methods, and implementation details.

\subsection{Dataset and Evaluation Metric}
Following the previous work~\cite{barba2020mulan}, we evaluated the systems with the updated version of SemEval-13 and SemEval-15 \footnote{https://github.com/SapienzaNLP/mwsd-datasets}(WordNet split).
Specifically, SemEval-13 contains four low-resourced languages: Italian (IT), Spanish (ES), French (FR), and German (DE), and SemEval-15 contains Italian and Spanish.
As no development dataset is provided, we randomly sample a small amount of the test instances as a development set for model selection. 
The instance number and the distribution of word sense number(\#sense) on the test dataset are shown in Table~\ref{tab:testset_statistics}. The word \#sense distribution is calculated separately on word level and instance level. The test instance with the higher word \#sense is more difficult than those with lower word \#sense because there are more senses to be disambiguated.
The F1 score(\%) is used as the evaluation metric.

\begin{table}
\small
\centering
\begin{tabular}{lccc}
\toprule
\textbf{Dataset} & \textbf{Inst num} & \textbf{Word avg} & \textbf{Inst avg} \\
\toprule
SemEval-13-IT & 1,490 & 3.80 & 5.51 \\
SemEval-13-ES & 1,260 & 4.20 & 5.52 \\
SemEval-13-FR & 1,449 & 2.36 & 3.03 \\
SemEval-13-DE & 1,076 & 1.60 & 2.17 \\
SemEval-15-IT & 1,007 & 4.38 & 5.27 \\
SemEval-15-ES & 1,043 & 6.17 & 6.19 \\
\bottomrule
\end{tabular}

\caption{Distribution of instance numbers, average word \#sense on word level, and average word \#sense on instance level for SemEval-13 and SemEval-15 test datasets. }\label{tab:testset_statistics}
\end{table}

\subsection{Training Corpora}
We utilize two types of automatically generated training datasets in our experiments, the dataset generated by our proposed translation-based method, and the dataset generated by a label propagation method MuLaN \cite{barba2020mulan}. Details of the dataset are shown in Table~\ref{tabel:translation number}. \\

\noindent \textbf{Translated Corpora:} SemCor \cite{miller1993semantic} is one of the largest annotated English WSD datasets, which contains 226,036 training instances covering 33,362 senses. We use SemCor 3.0 as the translation source. Due to differences in morphology between languages and inaccuracy brought by the alignment tool, a small amount of the annotated senses in English cannot be transferred to other languages. As a result, we get a comparable number of training instances. \\
    
\noindent \textbf{MuLaN:} MuLaN is one of the most representative works in automatically constructing the training corpus for the MWSD task. MuLaN has a broader coverage of sense keys as it utilizes the BabelNet inventory than SemCor which utilizes the WordNet inventory. Since we utilize the WordNet split of inventories and evaluation datasets, some of the words in the original MuLaN dataset are not in the used inventories. 
For fair comparison on mBERT-UNI model by inducing lexical gloss knowledge, we keep the instances with target words existing in the provided inventory, resulting in a filtered dataset MuLaN*.

\begin{table}
\small
    \centering
    \begin{tabular}{lccccc}
    \toprule
    \textbf{Language} & \textbf{EN} & \textbf{DE} & \textbf{FR} & \textbf{IT} & \textbf{ES} \\
    \midrule
    Translated & 226k & 169k & 181k & 181k & 179k \\
    MuLaN & -- & 245k & 311k & 416k & 452k \\
    MuLaN* & -- & 221k & 270k & 343k & 394k \\
    \bottomrule
    \end{tabular}
    \caption{Number of training instances for our translation based dataset, original MuLaN dataset, and filtered MuLaN* dataset.}
    \label{tabel:translation number}
\end{table}
\begin{table*}
\small
\centering
\begin{tabular}{lccccc c ccccc}
\toprule
 \multirow{2}{*}{Model}& \multicolumn{4}{c}{SemEval-13} && \multicolumn{2}{c}{SemEval-15} \\
\cmidrule{2-5} \cmidrule{7-8}
 & IT  & ES  & FR & DE  &&  IT  & ES \\
\midrule
BabelNet S1 & 53.22 & 60.32 & 60.04 & 76.58 && 45.38 & 39.31 \\
\midrule
mBERT-CLS (Trans) & 69.53 & \textbf{70.32} & 67.43 & 67.57 && 61.67 & 57.62 \\
\midrule
mBERT-UNI (SemCor) & 65.70 & 67.14 & \textbf{78.61} & 79.74 && 67.23 & 64.91 \\
mBERT-UNI (Trans) & \textbf{70.94} & 69.68 & 77.29 & \textbf{80.48} &&  \textbf{71.00} & \textbf{67.11} \\
\bottomrule
\end{tabular}
\caption{Results of mBERT-UNI on SemEval-13 and SemEval-15 test datasets. The training corpora generated from translation are briefly denoted as Trans.}
\label{tab:main_result}
\end{table*}
\subsection{Baselines}

We compare the proposed mBERT-UNI model with the following baseline methods:

\begin{enumerate}[leftmargin=*]
    \item \textbf{BabelNet S1}: This baseline tags the target word with its most common sense. Following the ranking in BabelNet inventory, the top one ranked sense is the most common sense (MCS). The left senses are least common sense (LCS).
    We denote this frequency-based baseline as ``BabelNet S1.''
    
    
    \item \textbf{mBERT-CLS}: The model is built on mBERT~\cite{kenton2019bert}. The pre-trained language model first extracts feature representation for target words in context sentences. On top of the frozen mBERT representation, a linear classifier is trained to classify the senses of target words. The model cannot be used to disambiguate unseen senses from the training dataset. Therefore, the model always predicts the most common sense for unseen senses as a back-off strategy.

\end{enumerate}

\subsection{Implementation Details}

Both encoders in mBERT-CLS and mBERT-UNI models are initialized with a pre-trained Bert-base-multilingual-uncased model, which has 110M parameters. For both models, we use the Cross-Entropy loss as the training loss, and Adam \cite{kingma2015adam} as the optimization algorithm.

For mBERT-CLS,  we fed the concatenation of the last four layers' output from mBERT encoder to a linear classifier. As discussed in \cite{blevins2020moving}, finetuning the mBERT-CLS does not improve the performance on the English WSD classification task. Therefore, we keep mBERT frozen and only train the linear classifier during training. The model is trained with a fixed learning rate $2 \cdot 10^{-5}$ for 50 epochs. The training batch size is 128.

For mBERT-UNI, the unified representation are generated from gloss knowledge in English, collecting from BabelNet and WordNet. For each sense key, BabelNet may have several gloss definitions and we select the source from WordNet for simplicity. The whole model is trained with the learning rate $10^{-5}$ for 20 epochs. We set the batch size at 40. The experiments are run on RTX 2080 and the average running time for each experiment is 40 hours. For collecting glosses of word senses, we use BabelNet API \footnote{https://babelnet.org/guide}.

%% file: files/6-results.tex
\section{Result Analysis}

\begin{table*}
\small
\centering
\begin{tabular}{lccccc c ccccc}
\toprule
\multirow{2}{*}{Model} & \multicolumn{4}{c}{SemEval-13} && \multicolumn{2}{c}{SemEval-15} \\
\cmidrule{2-5} \cmidrule{7-8}
     & IT  & ES  & FR & DE  &&  IT  & ES \\
\midrule
BabelNet S1 & 53.22 & 60.32 & 60.04 & 76.58 && 45.38 & 39.31 \\
\midrule
SensEmBERT \cite{scarlini2020sensembert} & 69.80 & 73.40 & 77.80 & 79.20 && - & - \\
OneSeC \cite{scarlini2019just} & 63.45 & 61.59 & 65.10 & 75.84 && - & - \\
MuLaN \cite{barba2020mulan} & \textbf{77.45} & 77.70 & 80.12 & 82.09 && 70.31 & 68.73 \\


mBERT-CLS (MuLaN*) & 69.73 & 75.87 & 78.54 & 82.62 && 68.82 & 67.50 \\
\midrule
mBERT-UNI (MuLaN*) & 75.64 & \textbf{80.24} & 81.64 & 83.27 && 72.99 & \textbf{70.47} \\
mBERT-UNI (Trans+MuLaN*) & 76.98 & 79.44 & \textbf{82.68} & \textbf{83.83} &&  \textbf{74.58} & 68.94 \\
\bottomrule
\end{tabular}
\caption{Results of mBERT-UNI with extra data corpora MuLaN on SemEval-13 and SemEval-15 test dataset. mBERT-CLS (MuLaN*) is the performance on filtered dataset MuLaN*.  mBERT-CLS (MuLaN*) is the performance of our implementation with filtered MuLaN as training data. MuLaN is the performance from original paper.}
\label{tab:mulan}\vspace{-1em}
\end{table*}

In this section, we analyze the performances of our proposed mBERT-UNI model in two parts. We first introduce the effects of mBERT-UNI on MWSD task with our generated translated corpora. After that, we present further experiment results on the MWSD task under various settings.

\subsection{Results of mBERT-UNI}\label{sec:result_main}

We present the performance of mBERT-UNI and other baseline methods in Table~\ref{tab:main_result}.
From the results, we can make the following observations:
\begin{enumerate}[leftmargin=*]
    \item Compared with BabelNet S1, knowledge and learning-based methods (i.e., mBERT-CLS and mBERT-UNI) can perform better in most languages. Such results show that even though we do not have any annotations for these languages, the corpus we translate from English can serve as a strong weak-supervision signal.
    \item The only exception is German, in which BabelNet S1 outperforms mBERT-CLS with translation. As shown in Table~\ref{tab:testset_statistics}, this is potentially because words in German typically have much fewer candidate senses than in other languages. As a result, in most instances, simply predicting the most common sense will lead to the correct answer. 
    In this case, the effect of learning is not as
    significant as in other languages.
    Even so, by carefully modeling the unified sense representations, the proposed model can still outperform the BabelNet S1 method by a 3.9 \% F1 score.
    \item Compared with the mBERT-CLS system, the proposed mBERT-UNI model outperforms on five out of six datasets because of additional lexical knowledge from sense representations with the same translated corpora. Though mBERT-CLS has captured the transferred supervised signal from translated corpora, it is still not enough to disambiguate the senses well. By utilizing lexical knowledge from the unified sense definitions, mBERT-UNI can better disambiguate the word senses under a supervised setting.
    \item The translated corpora benefit the MWSD system with external multilingual data. Compared to the mBERT-UNI system trained on original English SemCor and trained on the translated corpora, we can find that the system achieves performance gain on five out of six test datasets. This shows that though the machine translation and alignment tools may induce noise in the corpora preparation process, the resulting multilingual corpora still benefits the system on MWSD tasks. Future work may exploit in the direction of acquiring multilingual corpora of higher quality through automatic methods that can still benefit the system.
\end{enumerate}

\subsection{Further Analysis on mBERT-UNI}

In this section, we conduct further analysis to show the effects of leveraging an additional corpus MulaN~\cite{barba2020mulan} on mBERT-UNI, the effects of joint learning, and the performance on Least Common Sense (LCS). Details are as follows. 

\subsubsection{Effect of Adding MuLaN Corpora} \label{sec:result_mulan}

To see if the knowledge brought by the unified sense representations can be helpful under a supervised paradigm with extra training corpora, we conduct experiments on MuLaN. The results are shown in Table \ref{tab:mulan}.

By incorporating the unified sense representation, previous data generation methods such as MuLaN can further boost the performance of MWSD tasks. From the results, we can see even with the filtered training corpora of smaller size, mBERT-UNI still achieves performance gain over five out of six test datasets compared to MuLaN \cite{barba2020mulan}. Though the unified sense representations are built based on glosses from the English language only, it can still benefit multiple languages since words share a set of synsets. Future research may continue to find if enriching the sense representations with resources from different languages would still benefit the system.

Moreover, the unified sense representations encoded with the gloss knowledge from BabelNet, are of high quality. SensEmBERT \cite{scarlini2020sensembert} produced BERT-based sense embeddings by exploiting mostly the semantic relations in BabelNet and Wikipedia for multiple languages separately. Compared with SensEmBERT, our unified sense representation can be simply acquired from the single lexical knowledge source WordNet and even achieves higher performance on the MWSD task. 

The mBERT-UNI also supports merging multiple sources of data generation efforts. Combining MuLaN* with our generated dataset, mBERT-UNI can boost the performance on four out of six test datasets. The only exception is Spanish (ES). As shown in Table \ref{tab:testset_statistics}, the test instances in ES are more challenging than in other languages, and thus they are potentially more vulnerable to the noise in the automatically generated training corpora. 

\begin{table*}[t]
\centering
\small
\begin{tabular}{lccccc c ccccc}
\toprule
\multirow{2}{*}{Model} & \multicolumn{4}{c}{SemEval-13} && \multicolumn{2}{c}{SemEval-15} \\
\cmidrule{2-5} \cmidrule{7-8}
     & IT  & ES  & FR & DE  &&  IT  & ES \\
\midrule
mBERT-CLS (Trans) & 60.02 & \underline{63.11} & 57.16 & 45.58 && 49.26 & 53.21 \\
mBERT-UNI (Trans) & \underline{62.59} & 61.92 & \underline{67.38} & \underline{59.83} && \underline{63.90} & \underline{64.15} \\
\midrule

mBERT-CLS (MuLaN*) & 61.45 & 68.74 & 68.36 & 64.38 && 59.02 & \underline{63.61} \\
mBERT-UNI (MuLaN*) & \underline{68.24} & \underline{75.81} & \underline{72.77} & \underline{66.38} && \underline{66.83} & 63.47 \\

\bottomrule
\label{sec:result_LFS}\vspace{-1.5em}
\end{tabular}

\caption{Results of Least Common Senses (LCS) on SemEval-13 and SemEval-15 test dataset.}\label{tab:lfs}
\end{table*}

\subsubsection{Effect of Joint Training} \label{sec:result_joint}

\begin{figure}[t]
\centering
\includegraphics[clip, trim=1.8cm 1.0cm 2.2cm 2.0cm, width=\linewidth]{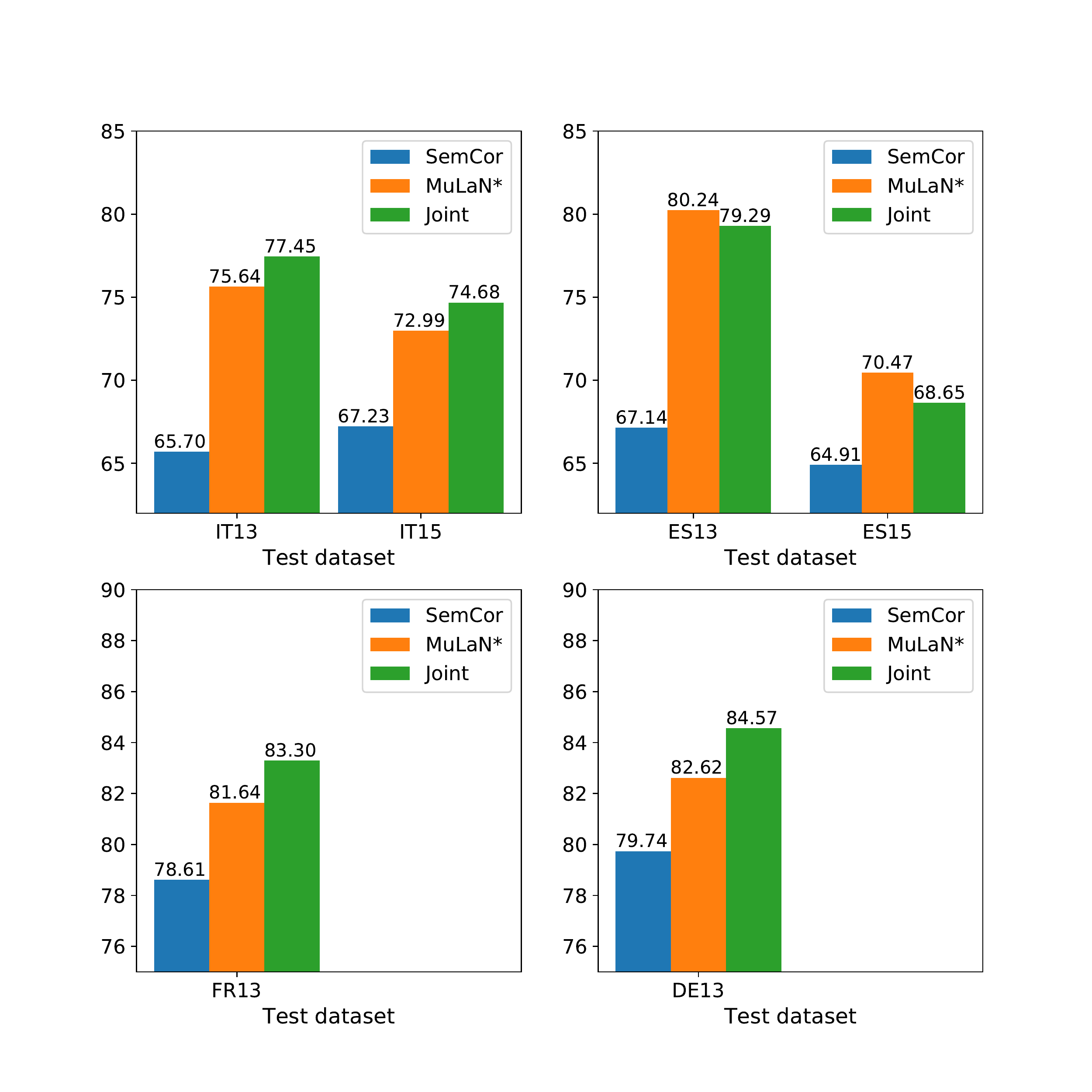}
\caption{Results of joint learning on the MWSD task. For joint training setting of each language, training data contains SemCor (English) and MuLaN* with the corresponding language part, e.g., SemCor and MuLaN* (Italian) for Italian (\textit{IT13} and \textit{IT15}).}
\label{fig:effect_on_other}
\end{figure}

\begin{table}[t]
\centering
\small
\begin{tabular}{lccccc}
\toprule
Dataset & SemCor & +IT & +ES & +FR & +DE\\
\midrule
ALL  & 75.70 & 75.94 &  75.82 & 75.76 & 75.31\\
$\Delta$ & - & +0.24 & +0.12 & +0.06 & -0.39\\
\bottomrule
\end{tabular}
\caption{Results of joint learning on the ALL test dataset of English WSD task. \textit{+IT} means that training dataset contains the SemCor (English) and MuLaN* (Italian).  }
\label{tab:ret_on_WSD}
\vspace{-1em}
\end{table}

In this section, we conduct experiments to study the effect of the proposed joint learning setting. We are interested in two questions: (1) Whether the joint learning setting can help models solve the MWSD problem or not? (2) Whether the joint learning setting will have a negative effect on English WSD or not.
To answer the question, we conduct experiments on training with monolingual datasets and multilingual datasets.

To answer the first question, we present the performances on the MWSD task in Figure~\ref{fig:effect_on_other}. 
We can see that joint learning can achieve better performance on four out of the six datasets and comparable performance on the other two (ES13, ES15). For each language, we combine the MuLaN* with English SemCor as a new training dataset. This result shows that with the unified sense representation, jointly training instances from different languages can improve the annotation usage efficiency across languages. For language ES, the higher word \#sense of the test instances may account for the performance drop under a joint training setting than using MuLaN* alone. 

To answer the second question, we report the performance of the mBERT-UNI model trained with English SemCor as well as another joint trained setting on the all-words English WSD datasets proposed by \cite{raganato2017word}. The test dataset ``ALL'' covers all five datasets, including semeval 2007 \cite{pradhan2007semeval}, senseval-2 \cite{palmer2001english}, senseval-3 \cite{snyder2004english}, semeval2012 \cite{navigli2013semeval}, and semeval2015 \cite{moro2015semeval}.
We show the results in Table~\ref{tab:ret_on_WSD}. We can see that the overall performance in English is comparable in different settings. Since the MuLaN dataset is specially designed for other languages and the propagated annotations mainly come from SemCor, the joint training does not benefit the English WSD task much. However, joint training enables a single mBERT-UNI to generate unified sense representations, which can be used in disambiguating word senses in multiple languages. In future work, the unified sense representation may be applied in cross-lingual representation learning.

\subsubsection{LCS Analysis} 
In this section, we analyze the influence of unified sense representation on the performance of the least common senses (LCS). We split the test instances into two parts, one part with annotation of BabelNet S1 and one part with annotations except BabelNet S1 as least common senses. Compared with most common senses, less common ones are more difficult to disambiguate for MWSD systems because of fewer training instances on average. 

We show the performances on the two groups of different systems are shown in Table~\ref{tab:lfs}. Comparing mBERT-CLS and mBERT-UNI, adding the sense representations can help improve models' performance significantly on the least common senses. The improvement is consistent on different training corpora for both the translated corpora and MuLaN. It can be concluded that unified sense representation with lexical knowledge improves the ability of deep models to disambiguate the least common senses. This is because mBERT-UNI can still generate and learn unique sense representations for the least common senses even with no or very few training instances.
However, while the systems achieve decent performance on the overall performance, disambiguating the lease common senses is still a challenging problem. 




%% file: files/7-conclusion.tex
\section{Conclusion and Future Work}

In this paper, to build feasible knowledge and supervised based systems for multilingual word sense disambiguation, we propose to construct unified sense representation by utilizing Babel synsets, and transferred annotations from rich source languages by machine translation and alignment tools. With the unified representations, previous data generation efforts can be combined and further boost the performance. Moreover, annotations from different languages can be jointly trained and benefit the multilingual word sense disambiguation task. Experiments on standard evaluation multilingual word sense disambiguation benchmarks demonstrate the effectiveness of the proposed method.

Future work can be extended on how to induce more lexical knowledge from various languages to improve the representation learning. 
Moreover, based on the fact that multiple languages share a set of concepts described by Babel synsets, the generated representations may benefit cross lingual representation for other natural language understanding tasks. 

%% file: acl.bbl
\begin{thebibliography}{40}
\expandafter\ifx\csname natexlab\endcsname\relax\def\natexlab#1{#1}\fi

\bibitem[{Banerjee et~al.(2003)Banerjee, Pedersen
  et~al.}]{banerjee2003extended}
Satanjeev Banerjee, Ted Pedersen, et~al. 2003.
\newblock Extended gloss overlaps as a measure of semantic relatedness.
\newblock In \emph{Ijcai}, volume~3, pages 805--810. Citeseer.

\bibitem[{Barba et~al.(2020)Barba, Procopio, Campolungo, Pasini, and
  Navigli}]{barba2020mulan}
Edoardo Barba, Luigi Procopio, Niccolo Campolungo, Tommaso Pasini, and Roberto
  Navigli. 2020.
\newblock Mulan: Multilingual label propagation for word sense disambiguation.
\newblock In \emph{Proc. of IJCAI}, pages 3837--3844.

\bibitem[{Bevilacqua and Navigli(2020)}]{bevilacqua2020breaking}
Michele Bevilacqua and Roberto Navigli. 2020.
\newblock Breaking through the 80\% glass ceiling: Raising the state of the art
  in word sense disambiguation by incorporating knowledge graph information.
\newblock In \emph{Proceedings of the 58th Annual Meeting of the Association
  for Computational Linguistics}, pages 2854--2864.

\bibitem[{Bevilacqua et~al.(2021)Bevilacqua, Pasini, Raganato, Navigli
  et~al.}]{bevilacqua2021recent}
Michele Bevilacqua, Tommaso Pasini, Alessandro Raganato, Roberto Navigli,
  et~al. 2021.
\newblock Recent trends in word sense disambiguation: A survey.
\newblock In \emph{Proceedings of the Thirtieth International Joint Conference
  on Artificial Intelligence, IJCAI-21}. International Joint Conference on
  Artificial Intelligence, Inc.

\bibitem[{Blevins and Zettlemoyer(2020)}]{blevins2020moving}
Terra Blevins and Luke Zettlemoyer. 2020.
\newblock Moving down the long tail of word sense disambiguation with gloss
  informed bi-encoders.
\newblock In \emph{ACL}.

\bibitem[{Bovi et~al.(2015)Bovi, Anke, and Navigli}]{bovi2015knowledge}
Claudio~Delli Bovi, Luis~Espinosa Anke, and Roberto Navigli. 2015.
\newblock Knowledge base unification via sense embeddings and disambiguation.
\newblock In \emph{Proceedings of the 2015 Conference on Empirical Methods in
  Natural Language Processing}, pages 726--736.

\bibitem[{Dou and Neubig(2021)}]{dou2021word}
Zi-Yi Dou and Graham Neubig. 2021.
\newblock Word alignment by fine-tuning embeddings on parallel corpora.
\newblock In \emph{Proceedings of the 16th Conference of the European Chapter
  of the Association for Computational Linguistics: Main Volume}, pages
  2112--2128.

\bibitem[{Dyer et~al.(2013)Dyer, Chahuneau, and Smith}]{dyer2013simple}
Chris Dyer, Victor Chahuneau, and Noah~A Smith. 2013.
\newblock A simple, fast, and effective reparameterization of ibm model 2.
\newblock In \emph{Proceedings of the 2013 Conference of the North American
  Chapter of the Association for Computational Linguistics: Human Language
  Technologies}, pages 644--648.

\bibitem[{Eisele and Chen(2010)}]{eisele2010multiun}
Andreas Eisele and Yu~Chen. 2010.
\newblock Multiun: A multilingual corpus from united nation documents.
\newblock In \emph{LREC}.

\bibitem[{Gale et~al.(1992)Gale, Church, and Yarowsky}]{gale1992method}
William~A Gale, Kenneth~W Church, and David Yarowsky. 1992.
\newblock A method for disambiguating word senses in a large corpus.
\newblock \emph{Computers and the Humanities}, 26(5):415--439.

\bibitem[{Hauer et~al.(2021)Hauer, Kondrak, Luan, Mallik, and
  Mou}]{hauer2021semi}
Bradley Hauer, Grzegorz Kondrak, Yixing Luan, Arnob Mallik, and Lili Mou. 2021.
\newblock Semi-supervised and unsupervised sense annotation via translations.
\newblock In \emph{Proceedings of the International Conference on Recent
  Advances in Natural Language Processing (RANLP 2021)}, pages 504--513.

\bibitem[{Huang et~al.(2019)Huang, Sun, Qiu, and Huang}]{huang2019glossbert}
Luyao Huang, Chi Sun, Xipeng Qiu, and Xuan-Jing Huang. 2019.
\newblock Glossbert: Bert for word sense disambiguation with gloss knowledge.
\newblock In \emph{Proceedings of the 2019 Conference on Empirical Methods in
  Natural Language Processing and the 9th International Joint Conference on
  Natural Language Processing (EMNLP-IJCNLP)}, pages 3509--3514.

\bibitem[{Iacobacci et~al.(2016)Iacobacci, Pilehvar, and
  Navigli}]{iacobacci2016embeddings}
Ignacio Iacobacci, Mohammad~Taher Pilehvar, and Roberto Navigli. 2016.
\newblock Embeddings for word sense disambiguation: An evaluation study.
\newblock In \emph{Proceedings of the 54th Annual Meeting of the Association
  for Computational Linguistics (Volume 1: Long Papers)}, pages 897--907.

\bibitem[{Kenton and Toutanova(2019)}]{kenton2019bert}
Jacob Devlin Ming-Wei~Chang Kenton and Lee~Kristina Toutanova. 2019.
\newblock Bert: Pre-training of deep bidirectional transformers for language
  understanding.
\newblock In \emph{Proceedings of NAACL-HLT}, pages 4171--4186.

\bibitem[{Kingma and Ba(2015)}]{kingma2015adam}
Diederik~P Kingma and Jimmy Ba. 2015.
\newblock Adam: A method for stochastic optimization.
\newblock In \emph{ICLR (Poster)}.

\bibitem[{Kumar et~al.(2019)Kumar, Jat, Saxena, and Talukdar}]{kumar2019zero}
Sawan Kumar, Sharmistha Jat, Karan Saxena, and Partha Talukdar. 2019.
\newblock Zero-shot word sense disambiguation using sense definition
  embeddings.
\newblock In \emph{Proceedings of the 57th Annual Meeting of the Association
  for Computational Linguistics}, pages 5670--5681.

\bibitem[{Langone et~al.(2004)Langone, Haskell, and
  Miller}]{langone2004annotating}
Helen Langone, Benjamin~R Haskell, and George~A Miller. 2004.
\newblock Annotating wordnet.
\newblock In \emph{Proceedings of the Workshop Frontiers in Corpus Annotation
  at HLT-NAACL 2004}, pages 63--69.

\bibitem[{Liu et~al.(2018)Liu, Lu, and Neubig}]{liu2018handling}
Frederick Liu, Han Lu, and Graham Neubig. 2018.
\newblock Handling homographs in neural machine translation.
\newblock In \emph{Proceedings of the 2018 Conference of the North American
  Chapter of the Association for Computational Linguistics: Human Language
  Technologies, Volume 1 (Long Papers)}, pages 1336--1345.

\bibitem[{Loureiro and Jorge(2019)}]{loureiro2019language}
Daniel Loureiro and Alipio Jorge. 2019.
\newblock Language modelling makes sense: Propagating representations through
  wordnet for full-coverage word sense disambiguation.
\newblock In \emph{Proceedings of the 57th Annual Meeting of the Association
  for Computational Linguistics}, pages 5682--5691.

\bibitem[{Luan et~al.(2020)Luan, Hauer, Mou, and Kondrak}]{luan2020improving}
Yixing Luan, Bradley Hauer, Lili Mou, and Grzegorz Kondrak. 2020.
\newblock Improving word sense disambiguation with translations.
\newblock In \emph{Proceedings of the 2020 Conference on Empirical Methods in
  Natural Language Processing (EMNLP)}, pages 4055--4065.

\bibitem[{Luo et~al.(2018)Luo, Liu, Xia, Chang, and Sui}]{luo2018incorporating}
Fuli Luo, Tianyu Liu, Qiaolin Xia, Baobao Chang, and Zhifang Sui. 2018.
\newblock Incorporating glosses into neural word sense disambiguation.
\newblock In \emph{Proceedings of the 56th Annual Meeting of the Association
  for Computational Linguistics (Volume 1: Long Papers)}, pages 2473--2482.

\bibitem[{Miller(1998)}]{miller1998wordnet}
George~A Miller. 1998.
\newblock \emph{WordNet: An electronic lexical database}.
\newblock MIT press.

\bibitem[{Miller et~al.(1993)Miller, Leacock, Tengi, and
  Bunker}]{miller1993semantic}
George~A Miller, Claudia Leacock, Randee Tengi, and Ross~T Bunker. 1993.
\newblock A semantic concordance.
\newblock In \emph{HUMAN LANGUAGE TECHNOLOGY: Proceedings of a Workshop Held at
  Plainsboro, New Jersey, March 21-24, 1993}.

\bibitem[{Moro and Navigli(2015)}]{moro2015semeval}
Andrea Moro and Roberto Navigli. 2015.
\newblock Semeval-2015 task 13: Multilingual all-words sense disambiguation and
  entity linking.
\newblock In \emph{Proceedings of the 9th international workshop on semantic
  evaluation (SemEval 2015)}, pages 288--297.

\bibitem[{Navigli et~al.(2013)Navigli, Jurgens, and
  Vannella}]{navigli2013semeval}
Roberto Navigli, David Jurgens, and Daniele Vannella. 2013.
\newblock Semeval-2013 task 12: Multilingual word sense disambiguation.
\newblock In \emph{Second Joint Conference on Lexical and Computational
  Semantics (* SEM), Volume 2: Proceedings of the Seventh International
  Workshop on Semantic Evaluation (SemEval 2013)}, pages 222--231.

\bibitem[{Navigli and Ponzetto(2012)}]{NAVIGLI2012217}
Roberto Navigli and Simone~Paolo Ponzetto. 2012.
\newblock \href {https://doi.org/https://doi.org/10.1016/j.artint.2012.07.001}
  {Babelnet: The automatic construction, evaluation and application of a
  wide-coverage multilingual semantic network}.
\newblock \emph{Artificial Intelligence}, 193:217 -- 250.

\bibitem[{{\"O}stling and Tiedemann(2016)}]{ostling2016efficient}
Robert {\"O}stling and J{\"o}rg Tiedemann. 2016.
\newblock Efficient word alignment with markov chain monte carlo.
\newblock \emph{The Prague Bulletin of Mathematical Linguistics}.

\bibitem[{Palmer et~al.(2001)Palmer, Fellbaum, Cotton, Delfs, and
  Dang}]{palmer2001english}
Martha Palmer, Christiane Fellbaum, Scott Cotton, Lauren Delfs, and Hoa~Trang
  Dang. 2001.
\newblock English tasks: All-words and verb lexical sample.
\newblock In \emph{Proceedings of SENSEVAL-2 Second International Workshop on
  Evaluating Word Sense Disambiguation Systems}, pages 21--24.

\bibitem[{Pasini(2020)}]{pasini2020knowledge}
Tommaso Pasini. 2020.
\newblock The knowledge acquisition bottleneck problem in multilingual word
  sense disambiguation.
\newblock In \emph{IJCAI}, pages 4936--4942.

\bibitem[{Pasini and Navigli(2020)}]{pasini2020train}
Tommaso Pasini and Roberto Navigli. 2020.
\newblock Train-o-matic: Supervised word sense disambiguation with no (manual)
  effort.
\newblock \emph{Artificial Intelligence}, 279:103215.

\bibitem[{Pasini et~al.(2021)Pasini, Raganato, Navigli et~al.}]{pasini2021xl}
Tommaso Pasini, Alessandro Raganato, Roberto Navigli, et~al. 2021.
\newblock Xl-wsd: An extra-large and cross-lingual evaluation framework for
  word sense disambiguation.
\newblock In \emph{Proceedings of the AAAI Conference on Artificial
  Intelligence}. AAAI Press.

\bibitem[{Pradhan et~al.(2007)Pradhan, Loper, Dligach, and
  Palmer}]{pradhan2007semeval}
Sameer Pradhan, Edward Loper, Dmitriy Dligach, and Martha Palmer. 2007.
\newblock Semeval-2007 task-17: English lexical sample, srl and all words.
\newblock In \emph{Proceedings of the fourth international workshop on semantic
  evaluations (SemEval-2007)}, pages 87--92.

\bibitem[{Pu et~al.(2018)Pu, Pappas, Henderson, and
  Popescu-Belis}]{pu2018integrating}
Xiao Pu, Nikolaos Pappas, James Henderson, and Andrei Popescu-Belis. 2018.
\newblock Integrating weakly supervised word sense disambiguation into neural
  machine translation.
\newblock \emph{Transactions of the Association for Computational Linguistics},
  6:635--649.

\bibitem[{Raganato et~al.(2017)Raganato, Camacho-Collados, and
  Navigli}]{raganato2017word}
Alessandro Raganato, Jose Camacho-Collados, and Roberto Navigli. 2017.
\newblock Word sense disambiguation: A unified evaluation framework and
  empirical comparison.
\newblock In \emph{Proceedings of the 15th Conference of the European Chapter
  of the Association for Computational Linguistics: Volume 1, Long Papers},
  pages 99--110.

\bibitem[{Scarlini et~al.(2019)Scarlini, Pasini, and
  Navigli}]{scarlini2019just}
Bianca Scarlini, Tommaso Pasini, and Roberto Navigli. 2019.
\newblock Just “onesec” for producing multilingual sense-annotated data.
\newblock In \emph{Proceedings of the 57th Annual Meeting of the Association
  for Computational Linguistics}, pages 699--709.

\bibitem[{Scarlini et~al.(2020)Scarlini, Pasini, and
  Navigli}]{scarlini2020sensembert}
Bianca Scarlini, Tommaso Pasini, and Roberto Navigli. 2020.
\newblock Sensembert: Context-enhanced sense embeddings for multilingual word
  sense disambiguation.
\newblock In \emph{Proceedings of the AAAI Conference on Artificial
  Intelligence}, volume~34, pages 8758--8765.

\bibitem[{Snyder and Palmer(2004)}]{snyder2004english}
Benjamin Snyder and Martha Palmer. 2004.
\newblock The english all-words task.
\newblock In \emph{Proceedings of SENSEVAL-3, the Third International Workshop
  on the Evaluation of Systems for the Semantic Analysis of Text}, pages
  41--43.

\bibitem[{Taghipour and Ng(2015)}]{taghipour2015one}
Kaveh Taghipour and Hwee~Tou Ng. 2015.
\newblock One million sense-tagged instances for word sense disambiguation and
  induction.
\newblock In \emph{Proceedings of the nineteenth conference on computational
  natural language learning}, pages 338--344.

\bibitem[{Tiedemann et~al.(2020)Tiedemann, Thottingal
  et~al.}]{tiedemann2020opus}
J{\"o}rg Tiedemann, Santhosh Thottingal, et~al. 2020.
\newblock Opus-mt--building open translation services for the world.
\newblock In \emph{Proceedings of the 22nd Annual Conference of the European
  Association for Machine Translation}. European Association for Machine
  Translation.

\bibitem[{Wu et~al.(2016)Wu, Schuster, Chen, Le, Norouzi, Macherey, Krikun,
  Cao, Gao, Macherey et~al.}]{wu2016google}
Yonghui Wu, Mike Schuster, Zhifeng Chen, Quoc~V Le, Mohammad Norouzi, Wolfgang
  Macherey, Maxim Krikun, Yuan Cao, Qin Gao, Klaus Macherey, et~al. 2016.
\newblock Google's neural machine translation system: Bridging the gap between
  human and machine translation.
\newblock \emph{arXiv preprint arXiv:1609.08144}.

\end{thebibliography}
